# GuardNet: Ensemble Strategies of Shallow Neural Networks for Robust Prompt Injection and Jailbreak Detection

Paulo Ricardo Ferreira Neves[1], Edson Rodrigues da Cruz Filho[1,3], Paulo Henrique Eleuterio Falsetti[1,2], João Vitor Pavan[1], Ian Degaspari[1], Henrique Vieira Laturrague[1], Patrick Vieira Laturrague[1], Guilherme Nielsen Dias[1], Marccello Wilson Perez Berto[1], Gustavo Voltani Von Atzingen[1,3]

[1] Quickium Technology Ltd. – Piracicaba, São Paulo

[2] Federal University of São Carlos (UFSCar) – Sorocaba Campus – São Paulo

[3] Federal Institute of Education, Science and Technology of São Paulo (IFSP) – Piracicaba Campus – São Paulo

## Abstract

Large Language Models (LLMs) have transformed natural language processing, but they remain vulnerable to Prompt Injection (PI) and Jailbreak (JB) attacks. In addition, benchmark evaluations may be affected by contamination and partial information leakage, compromising performance estimates. This work presents GuardNet, a guardrail system based on an ensemble of shallow neural networks (BiLSTMs) with approximately 47 million parameters. We investigate the hypothesis that robustness in adversarial scenarios depends more on the diversity of example coverage and threshold calibration than on model scale. The results indicate that GuardNet achieves competitive performance compared with lightweight detectors and high efficiency at low latency, although larger LLMs such as Mistral-7B and Llama-3.1-8B still achieve superior performance in terms of F1 score and AUROC on the blind JBB-Behaviors benchmark. Nevertheless, GuardNet achieves an AUROC of 0.747 on the blind dataset (n = 200) and an F1 score of 0.92 on a proprietary benchmark (n = 50), under threshold calibration and evaluation with declared partial information leakage. The system operates with an average latency of approximately 50 ms on CPU, making it suitable for deployment in production environments with cost and infrastructure constraints.

## 1. Introduction

The modern concept of Large Language Models (LLMs) gained prominence with the development of scalable Transformer-based architectures, particularly following the release of GPT-3 by OpenAI (Brown et al., 2020), which demonstrated the ability to generate coherent, high-quality text across a wide range of natural language processing (NLP) tasks (Perez & Ribeiro, 2022). In general, the field of NLP encompasses tasks such as machine translation, text summarization, information extraction, and text classification, and has evolved significantly over the past decades (Khurana et al., 2023). The introduction of GPT-3 marked a new phase in natural language processing. However, important advances had already been achieved by models such as BERT (Devlin et al., 2019), which introduced deep bidirectional representations for language understanding. Subsequently, other large language models, such as GPT-J (Wang & Komatsuzaki, 2021), T5 (Raffel et al., 2020), and OPT (Zhang et al., 2022), further advanced the state of the art across a variety of NLP tasks (Perez & Ribeiro, 2022).

LLM-based applications can be developed using simple prompts with user-input substitution, enabling the execution of complex tasks that would be difficult to implement with traditional rule-based systems (Perez & Ribeiro, 2022). However, this ease of development also facilitates attacks through the injection of malicious instructions, a process known as prompt injection, making the protection of such applications challenging due to the open-ended nature of prompts (Perez & Ribeiro, 2022). In this context, prompt injection is defined as the insertion of malicious instructions into prompts with the objective of misaligning or manipulating the behavior of an LLM (Perez & Ribeiro, 2022).

Furthermore, although alignment mechanisms and safety filters are employed to restrict inappropriate behaviors in LLMs, these models remain vulnerable to adversarial inputs (Wei et al., 2023). Adversarial inputs consist of manipulated or carefully modified inputs designed to induce machine learning models to produce incorrect or undesired behaviors, even when the applied modifications are subtle or appear legitimate (Kurakin et al., 2016; Szegedy et al., 2013). This process can be characterized as a jailbreak attack, that is, an attack designed to circumvent safety and alignment mechanisms, inducing the model to generate responses that would otherwise be blocked (Wei et al., 2023).

Prompt injection and jailbreak are security challenges that arise under conditions of strong distribution shift and adversarial behavior, in which the model is exposed to inputs outside the distribution represented in its safety training data and may exhibit alignment failures (Szegedy et al., 2013).

However, existing evaluations often fail to adequately distinguish between genuine generalization and overfitting to benchmarks, particularly in scenarios susceptible to data contamination and training information leakage (Carlini et al., 2021; D'Amour et al., 2020). Furthermore, although LLMs are widely used as classifiers in a variety of applications, evidence suggests that their operational behavior and confidence calibration have not yet been systematically characterized (C. Guo et al., 2017; Kadavath et al., 2022).

To mitigate these security and behavioral issues, guardrail mechanisms for machine learning models have been proposed as techniques to control or constrain model behavior, including methods such as Reinforcement Learning from Human Feedback (RLHF), input and output filters, and response restriction policies, with the goal of preventing the generation of inappropriate or unsafe content (Ouyang et al., 2022).

Given the limitations discussed, including the vulnerability of LLMs to adversarial inputs under conditions of distribution shift, as well as the fragility of evaluations based on potentially contaminated benchmarks and the lack of systematic calibration analysis, it remains unclear which factors effectively determine the robustness of these models in adversarial scenarios. In this context, this work investigates the hypothesis that the robustness of LLMs is more strongly influenced by the diversity and coverage of adversarial examples and by the proper calibration of decision thresholds than by model scale or architecture.

## 2. Related Work

This section presents related work on the detection of Prompt Injection (PI) and Jailbreak (JB) attacks, with a focus on lightweight detectors based on classical machine learning algorithms, shallow neural networks, and compact encoders.

### 2.1 Shallow Neural Networks

Shallow neural networks are models composed of a small number of processing layers, typically a single hidden layer between the input and output layers (Goodfellow, 2016). Recent studies indicate that lightweight models, such as shallow neural networks, may exhibit greater robustness to obfuscation attacks. Kushnerov et al. (2026) demonstrated that architectures such as Word-BiLSTM (F1 = 0.9681) and CNN (F1 ≈ 0.962) outperform Transformers (F1 = 0.9190) on the Multi-Party Dialogue Dataset (MPDD), especially under obfuscation attacks. Notably, character-level (char-level) models, with approximately 22k to 24k parameters, appear to be the most robust against homograph and spacing attacks. Other works explore Siamese architectures, such as danSN (A. Guo et al., s.d.), which employs Siamese LSTMs as expert components within Mixture of Experts systems.

### 2.2 Classical Machine Learning and Embeddings

The use of traditional classifiers combined with pre-trained embeddings or simple text representations (TF-IDF) has proven to be highly effective. Corll (2026) introduced the “Mirror” pattern, arguing that strict data geometry outperforms model scale. Their linear SVM using character n-grams achieved an F1 score of 0.9207, significantly outperforming Meta’s Prompt-Guard-2 (F1 = 0.5914) on the same test set.

Recent industry results further reinforce the competitiveness of lightweight approaches for detecting attacks in LLMs. In a study conducted by NVIDIA, Galinkin and Sablotny (2024) demonstrated that a Random Forest classifier combined with Snowflake embeddings achieved an F1 score of 0.9601 on the JailbreakHub benchmark, while the PromptGuard model achieved only 0.3029 under the same conditions. Similarly, the MoJE framework developed by IBM (Cornacchia et al., 2024) employed an ensemble based on XGBoost and Logistic Regression to achieve an AUC of 0.9947, significantly outperforming Llama-Guard. These results suggest that classical machine learning models, when combined with appropriate text representations, can not only compete with Transformer-based architectures but also surpass them in specific scenarios of jailbreak and prompt injection detection.

### 2.3 Lightweight Encoders and Guardrail Mechanisms

Transformer-based models with smaller sizes, ranging from 67M to 184M parameters, currently act as intermediaries between classical machine learning classifiers and large language models (LLMs). Prominent examples in this category include deepset/deberta-v3, fmops/distilbert, and the Meta Prompt-Guard-2 family (*deepset/deberta-v3-base-injection · Hugging Face*, 2024; *fmops/distilbert-prompt-injection · Hugging Face*, 2024; *meta-llama/Llama-Prompt-Guard-2-86M · Hugging Face*, 2025). Although these models achieve high accuracy on in-distribution data and short-context scenarios, recent research reveals important limitations in their detection capabilities; for instance, Meta Llama Prompt Guard 2 (86M) experienced a detection collapse, resulting in a 100% bypass rate when facing Prompt Overflow attacks under an interleaved fragmentation layout (Y. Zhou et al., 2026). Zhou et al. (2026) show that this failure arises from a mismatch between the limited inspection windows of guardrails (often restricted to 512 tokens) and the large inference context windows of downstream LLMs, allowing malicious instructions distributed across the input to be locally ignored by the detector due to low signal density, while remaining fully reconstructible and actionable by the final model. Beyond context-inspection vulnerabilities, studies also highlight generalization limitations. The DMPI-PMHFE model, which uses a DeBERTa-v3 architecture integrated with heuristic feature engineering, exhibits high sensitivity to domain shifts, with its F1 score dropping from 0.98 to 0.90 across different datasets (Ji et al., 2025).

### 2.4 Benchmark Validity and Over-Defense

A central challenge highlighted in the literature is data contamination and overfitting in common benchmarks. The study *"When Benchmarks Lie"* (Fomin, 2026) warns that the AUC metric may be inflated by up to 8.4 points in standard data splits compared to Leave-One-Dataset-Out evaluations. Furthermore, detectors such as InjecGuard and PromptGuard often suffer from "over-defense", maintaining high recall for attack detection but collapsing in benign prompt classification, with accuracy rates dropping to between 5% and 35%. The comparative results of these models across different architectures and benchmarks are summarized in Table 1.

**Table 1.** Performance comparison between lightweight detectors and guardrail models based on different architectures, highlighting performance variations across datasets, parameter scales, and evaluation metrics, with a focus on attack and jailbreak detection scenarios.

| Model | Family / Architecture | Number of Parameters | Evaluation Dataset | Primary Metric |
|---|---|---|---|---|
| Char-BiLSTM (Kushnerov et al., 2026) | RNN (Char-level) | ~22 k | MPDD | F1 = 0,9599 |
| Char-CNN (Kushnerov et al., 2026) | CNN (Char-level) | ~23,9 k | MPDD | F1 = 0,9631 |
| Mirror (Corll, 2026) | SVM Linear | < 1 MB | Holdout-524 | F1 = 0,9207 |
| RF + Snowflake (Galinkin & Sablotny, 2024) | Random Forest | Árvore rasa | JailbreakHub | F1 = 0,9601 |
| Prompt-Guard-2 Small (*meta-llama/Llama-Prompt-Guard-2-86M · Hugging Face*, 2025) | Transformer | 22 M | Jailbreak (EN) | AUC = 0,995 |
| Prompt-Guard-2 Base *ama/Llama-Prompt-Guard-2-86M · Hugging Face*, 2025) | Transformer | 86 M | Jailbreak (EN) | AUC = 0,998 |

## 3. Methodology

This section describes the design of the GuardNet system, with a focus on the ensemble strategy for robust detection, data curation techniques that prioritize diversity over volume, and the implemented architecture.

### 3.1 Ensemble Design

GuardNet uses an ensemble architecture composed of three independent heads based on shallow BiLSTM neural networks (Bidirectional Long Short-Term Memory) (approximately 15M parameters each) (Graves & Schmidhuber, 2005; Schuster & Paliwal, 1997). The choice of shallow models (Goodfellow, 2016) is motivated by computational efficiency and the use of discriminative classifiers, which perform a direct mapping between textual inputs and supervised labels, without any objective of language modeling or autoregressive token generation (Domingos, 2012). Unlike LLMs, these models do not operate with decoding mechanisms (lm_head) and are designed exclusively for classification tasks, which reduces their exposure to prompt injection attacks based on instructions embedded in the input.

The outputs of the three ensemble heads are combined using an arithmetic mean of the predicted probabilities (Z.-H. Zhou, 2025). The final decision is obtained through a global threshold applied to the aggregated probability, empirically calibrated on the validation set ($\tau = 0.65$). This threshold is tuned to balance the behavior of the three sub-networks, reducing asymmetries between recall and precision observed in each individual head and promoting overall ensemble stability.

The identities and roles of the ensemble heads are presented in Table *2*. This complementary gating strategy enables the ensemble to achieve an F1 score of 0.92, outperforming any individual member.

**Table 2.** Roles of the neural network ensemble heads.

| Head | Function |
|---|---|
| GuardNet-v3 (Conservative Anchor) | Characterized by the highest standalone precision within the ensemble (FPR = 0.10), acting as the stability pillar to prevent false positives in production environments. |
| GuardNet-v9 (Aggressive Recall) | Trained with aggressive undersampling (1:1), this gate compensates for the gaps of v3 in subtle attacks, achieving 100% recall for jailbreak (JB) attacks on the reference benchmark. |
| GuardNet-T2 (Sanity Check) | A multi-class network (15 classes) that contributes to the ensemble as a secondary binary vote through the inverse probability of the benign class. |

### 3.2 Training Strategies: Adversarial Coverage vs. Scale

The training of GuardNet was based on the premise that the diversity of adversarial axes is a more critical factor than the raw volume of data. Accordingly, to ensure generalization and prevent overfitting to specific dataset distributions, four main protocols were applied: (i) dominance control, through the implementation of an absolute cap of 8,000 samples per source, where larger datasets are subsampled in a stratified manner and smaller datasets are included in full; (ii) stratification by attack axes, through the definition of minimum quotas for critical categories (persona-based jailbreaks (DAN), goal hijacking, data exfiltration via JSON, and multilingual attacks); (iii) false-positive robustness design, including benign samples with structural characteristics similar to attacks, such as short imperative prompts, requests for JSON-formatted outputs via function calling, and harmless role-play dialogues; (iv) source balancing, using a WeightedRandomSampler with weights ($w_i$) inversely proportional to the square root of the source size ($n_i$), as shown in Equation (1):

$$(1 \,/\, \sqrt{n_i}) \tag{01}$$

seeking a balance between uniform and proportional sampling.

### 3.3 Architecture for Model Deployment

Unlike defense approaches based on external services (Markov et al., 2023), GuardNet was implemented as an in-process component integrated into the Django framework (*Django*, s.d.). The models and tokenizers (based on bert-base-multilingual-cased) were embedded directly into the application using the PyTorch ("PyTorch", s.d.) and transformers (*Transformers · Hugging Face*, s.d.) libraries.

Checkpoint loading was implemented via lazy loading on the first request, enabling detection to occur synchronously in the critical path (hot path) of the user's HTTP request. This architecture is compatible with compact models optimized for CPU inference, reducing reliance on dedicated GPU infrastructure in low-latency and memory-constrained scenarios, as demonstrated by models such as DistilBERT (Sanh et al., 2019) and MobileBERT (Sun et al., 2020), and mitigating the serialization and network latency overhead inherent to communication between microservices (*Microservices*, s.d.).

The system was executed in an x86_64 environment with an Intel Core i7-13700F processor and an NVIDIA GeForce RTX 3060 GPU (12 GB VRAM). CPU inference was primarily used in the production pipeline, with optional GPU acceleration employed during testing and development stages.

## 4. Results

This section details the empirical performance of GuardNet-E, discussing the effectiveness of data curation, the phenomenon of contamination in external baselines, and the operational trade-offs between specialized models and general-purpose LLMs.

### 4.1 Evolution of Training Iterations: Diversity as a Driver of Generalization

The development of GuardNet was conducted as an exploratory analysis of the feasibility of the proposed solution, given the broad nature of the problem. Rather than a linear optimization process, multiple simultaneous challenges were observed, including linguistic diversity, diversity of attack techniques, subtle natural language injections, false positives in benign prompts with attack-like structure, and decision threshold calibration. Each iteration isolated and investigated one of these challenges, as detailed in Table 3.

The evaluation in this phase was conducted on awall-test, a synthetic and proprietary dataset generated by an LLM, consisting of 50 records (n = 50) covering benign prompts, prompt injection, and jailbreak cases. The use of a synthetic dataset was necessary because real-world labeled data are scarce, highly imbalanced, and often subject to non-commercial licenses, while public benchmarks also present contamination risks, as many models may have been exposed to them during pre-training. LLM-based generation enabled control over the attack taxonomy, the construction of a balanced and reproducible dataset free of licensing constraints, and the creation of novel variations to probe generalization. As this represents a feasibility probe rather than a confirmatory evaluation, the reduced sample size is appropriate: it provides signal and guides design decisions but does not support statistical significance claims. A partial and explicitly acknowledged leakage was present, since the generating LLM may reproduce publicly known attack patterns; therefore, a separate blind evaluation was conducted on the JBB-Behaviors benchmark with 200 records (n = 200).

**Table 3.** Outcome, decision, and justification of each iteration

| # | Model | Description / Decision | F1 awall-test (n = 50) | F1 JBB (n = 200) |
|---|---|---|---|---|
| 0 | GuardNet-v0 | The initial model of the project served as a proof of concept to assess feasibility. It learned patterns from the training data itself (high in-domain F1), but collapsed when evaluated on unseen data (deepset hold-out, F1 = 0.46), revealing clear overfitting. | n/a (proxy) | — |
| 1 | GuardNet-v1 | Expanded data coverage. Each new dataset targeted a specific challenge: linguistic diversity (multilingual_octavio), attack technique diversity (Smooth-3/llm-prompt-injection-attacks), and subtle natural language injections (wambosec/prompt-injections-subtle). | n/a (proxy) | — |
| 2 | GuardNet-v2 | Tested the opposite hypothesis: improving generalization through stronger regularization (L2 and dropout) without changing data. With the same pool as v1, gains were marginal (+0.048 F1), confirming that data diversity, not regularization alone, was the key factor. | n/a (proxy) | — |
| 3 | GuardNet-v3 | Expanded the dataset pool with a cap of 50k samples per source to prevent dominance of large datasets. BeaverTails revealed a structural weakness in harmful | 0.8718 | 0.667 |
| | | content detection: while other sources reached ~0.95 F1, BeaverTails remained at ~0.65, exposing a hard boundary for shallow models. | | |
| 4 | GuardNet-v4 | Further expansion with four adversarially rich datasets targeting recurring failure modes: real-world jailbreak personas (TrustAIRLab/in-the-wild), instruction override (gandalf), markup and trigger-token spoofing (HackAPrompt), and human red-teaming in natural dialogue (hh-rlhf). Achieved the highest pre-licensing performance (F1 = 0.7793). | 0.7391 | — |
| 5 | GuardNet-v5 | License audit revealed non-permissive sources. Retrained using only permissively licensed datasets (decision_pool). Resulting regression (~−0.16 F1) indicated that removed sources contained adversarial diversity not recovered by the remaining pool. | n/a (proxy) | — |
| 6 | GuardNet-v6 | Attempted recovery via increased epochs on the same restricted pool. Gains were marginal (+0.023 F1), confirming that additional training does not compensate for missing coverage. Motivated a shift away from purely training-time scaling. | n/a (proxy) | — |
| 7 | GuardNet-v7 | Checkpoint with uncertain provenance. Excluded from comparisons due to lack of reproducibility guarantees. | 0.7179 | — |
| 8 | GuardNet-v8 | Introduced hard negatives (benign instructions structurally similar to attacks) and shifted checkpoint selection toward recall optimization. | 0.8372 | — |
| 9 | GuardNet-v9 | Applied aggressive undersampling (1:1 balance). Precision improved (+0.034), but recall collapsed (−0.41), severely degrading subtle attack detection (F1 0.71 vs 0.95). However, it achieved 100% recall for jailbreak (JB) attacks and was later repurposed as a secondary gate in the ensemble. | 0.8085 | — |
| T1 | GuardNet-T1 | First multi-class attempt (14 attack types + benign) based on v3. Failed to generalize (macro F1 = 0.14), confirming that fine-grained attack taxonomy does not improve robustness in this setting. | — | — |
| T2 | GuardNet-T2 | Second multi-class attempt (15 classes). Performance remained similar to T1 (negligible AUROC gain +0.0017). Although ineffective as a standalone classifier, it serves as a binary voting signal via inverse benign probability (1 − P(benign)). | n/a (component) | — |
| E2 | GuardNet-E2 | First ensemble conFiguretion. High FPR (0.33), as the system inherited the over-sensitive behavior of v9 without a precision anchor. Discarded. | 0.7755 | — |
| E2′ | GuardNet-E2′ | Final ensemble candidate combining v3 (precision anchor), v9 (high-recall gate), and T2 (binary vote), aggregated via mean with threshold τ = 0.65. Achieved balanced performance (F1 = 0.85, FPR = 0.10), reducing false positives at some recall cost. | 0.8718 | 0.6735 |
| E | GuardNet-E (final) | Final ensemble conFiguretion combining v3, v9, and T2 with mean aggregation and threshold τ = 0.65. Achieves balanced behavior (F1 = 0.92, FPR = 0.10), resolving excessive false positives from | 0.9231 | 0.7143 |

| | | |
|---|---|---|
| | earlier ensembles while maintaining strong jailbreak detection capability. | |

In general, the development trajectory confirmed that adversarial diversity is more important than gains from parameter scale. In the initial iteration (GuardNet-v0), the model achieved high in-distribution performance; however, it exhibited severe failures on unseen data, with an F1 score of 0.46. Progressing to the next training stage, accompanied by an expansion of the training pool, GuardNet-v1 showed a substantial improvement on the hold-out set, reaching an F1 score of 0.87 without any architectural changes.

The gradual evolution of the system culminates in GuardNet-E, which achieves an F1 score of 0.92 on the validation set, highlighting the gains associated with broader coverage of adversarial axes. However, a significant sensitivity to decision threshold calibration is observed: performance varies from an F1 score of 0.77 ($\tau = 0.5$) to 0.92 ($\tau = 0.65$), indicating that a substantial portion of the performance depends on this hyperparameter tuning step. This behavior suggests that the ensemble exhibits a strong dependence on the validation distribution used for threshold optimization.

In blind evaluation on the JBB-Behaviors benchmark ($n = 200$), the model achieves an F1 score of 0.714, resulting in a generalization gap of approximately 0.206 relative to the calibrated validation performance. This discrepancy highlights the importance of decision threshold calibration in adversarial scenarios and suggests that apparent gains on non-blind sets may overestimate performance under truly unseen distributions.

The evolution shown in Figure 1 indicates that the largest performance gains did not result exclusively from changes in architecture, but primarily from modifications to the training dataset, data selection strategy, and model combination approach. The GuardNet-v3, GuardNet-v8, and GuardNet-E2′ versions achieved a relatively stable performance plateau between 0.80 and 0.87 maximum F1, suggesting that, beyond a certain point, improvements became marginal. In this context, GuardNet-E stands out as the best overall result, reaching approximately 0.92 F1 with a calibrated threshold of 0.65, which suggests that the ensemble strategy successfully balances high precision and high recall.

Another relevant aspect observed in the figure is that alternative architectures—including TextCNN, Transformer, and CNN-LSTM—remained concentrated within a similar performance range, between 0.70 and 0.80 F1. This behavior suggests that the neural network architecture, in isolation, was not the primary bottleneck of the investigated problem. Instead, the results indicate that factors such as training pool diversity, coverage of adversarial examples, balancing across data sources, and model combination strategies had a significantly greater impact on the system's generalization capability.

In addition, the stabilization of more advanced models within similar performance ranges reinforces the hypothesis of partial architectural saturation: beyond a certain level of complexity, increasing depth or changing network types yields smaller gains than those obtained through improvements in data quality and diversity. Thus, the figure supports the conclusion that, for adversarial content detection tasks, dataset engineering and ensemble design may be more influential than isolated architectural changes.

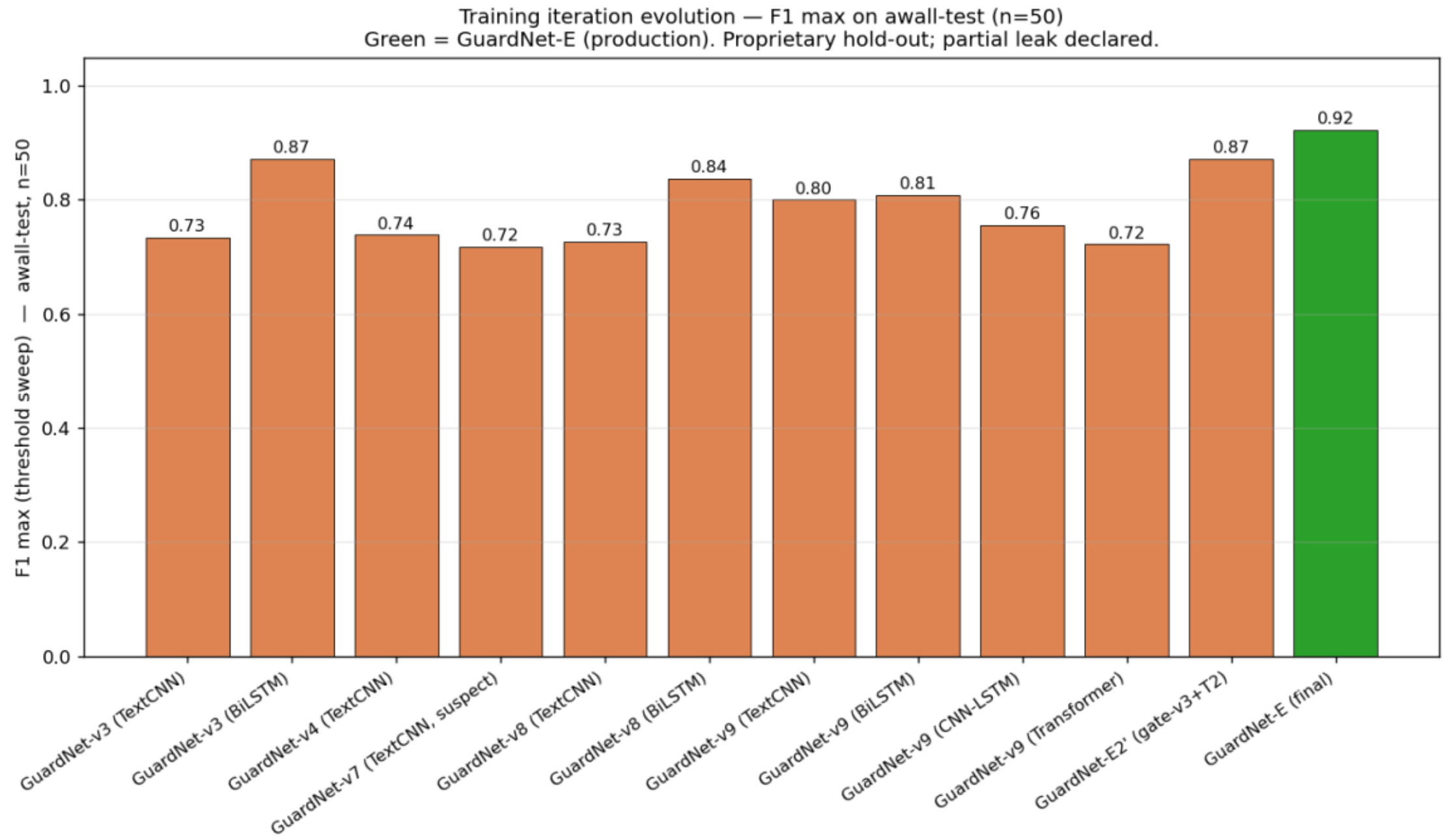

**Figure 1.** Chronological evolution of GuardNet model iterations. F1 scores correspond to the maximum values obtained through threshold sweeping on the proprietary benchmark (n = 50), which includes declared partial information leakage.

The analysis presented in Figure 2 shows that threshold calibration has a direct and, in some cases, decisive impact on the final performance of the classifiers. The comparison between F1 values obtained with a fixed threshold of 0.5 and the maximum values obtained through threshold sweeping indicates that a substantial portion of the model's capability remains "hidden" when only the default binary decision threshold is used. Thus, the difference between the blue and orange bars represents the performance gain recoverable solely through threshold tuning, without any architectural changes or additional model training.

The most pronounced case is GuardNet-E, whose performance increases from approximately 0.77 at $\tau = 0.5$ to around 0.92 at $\tau = 0.65$. This result shows that the ensemble produces informative probabilistic scores, but these are shifted relative to the default decision point. In other words, the model already exhibits strong discriminative capability, but the default threshold introduces an imbalance between precision and recall. Calibration allows this trade-off to be repositioned in a manner more consistent with the system's objectives.

These results also suggest that the threshold should be treated as an important hyperparameter in the inference pipeline, rather than a purely arbitrary post-processing choice. In security and adversarial detection applications, small changes in the threshold can significantly alter false positive and false negative rates, directly affecting practical usability. Therefore, calibration becomes an essential part of the experimental pipeline. However, the analysis also highlights an important methodological consideration: the threshold should not be tuned directly on the final test set, as this introduces data leakage and leads to overly optimistic performance estimates. Ideally, calibration should be performed on a separate validation or development set, preserving the test set as a truly blind measure of generalization. Thus, while the observed gains are substantial, they also reinforce the need for rigorous evaluation protocols to avoid overestimation of results.

Figure 2 provides a clear view of how the GuardNet system evolved from isolated models to an ensemble strategy of shallow neural networks, achieving strong performance on the awall-test benchmark. It is notable that GuardNet-E (final), represented by the thick green curve, dominates the plot with an AUC of 0.947, approaching the "ideal top-left corner" where recall is maximized and false positives are minimized.

Comparing the early curves, it becomes clear that architectural choices played a significant role. GuardNet-v3 (BiLSTM) achieves an AUC of 0.932, significantly outperforming GuardNet-v3 (TextCNN), which stabilizes at 0.847. This visually validates the design decision to use three independent BiLSTM-based heads (each with 15M parameters), leveraging their ability to capture sequential dependencies that are missed by purely convolutional models.

The dashed curves (v8 and v9) represent models operating under different trade-offs. GuardNet-v9, for instance, was trained with aggressive undersampling to ensure 100% recall for jailbreak attacks, but this comes at the cost of precision calibration. The transition to GuardNet-E demonstrates that combining these complementary "gates"—merging the precision anchor (v3) with the high-recall behavior of other components—yields a more robust global class separation.

The evolution shown in Figure 2 supports the central claim of this work: adversarial robustness depends more on coverage diversity and threshold calibration than on large-scale parameter growth. While models such as Prompt-Guard-2 (86M) are vulnerable to fragmentation attacks like Prompt Overflow, which can reach 100% bypass rates under interleaved layouts, GuardNet relies on a strict data geometry approach. The figure shows that, by adjusting the threshold (moving from 0.77 to 0.92 F1 via threshold sweep), the ensemble recovers discriminative capacity that general-purpose models often lose due to insufficient calibration.

In summary, Figure 2 is not merely a collection of metrics, but evidence that well-curated shallow neural networks can provide a deterministic and efficient security layer (approximately 50 ms CPU latency) that rivals or surpasses larger classifiers, provided they operate under a properly calibrated ensemble strategy.

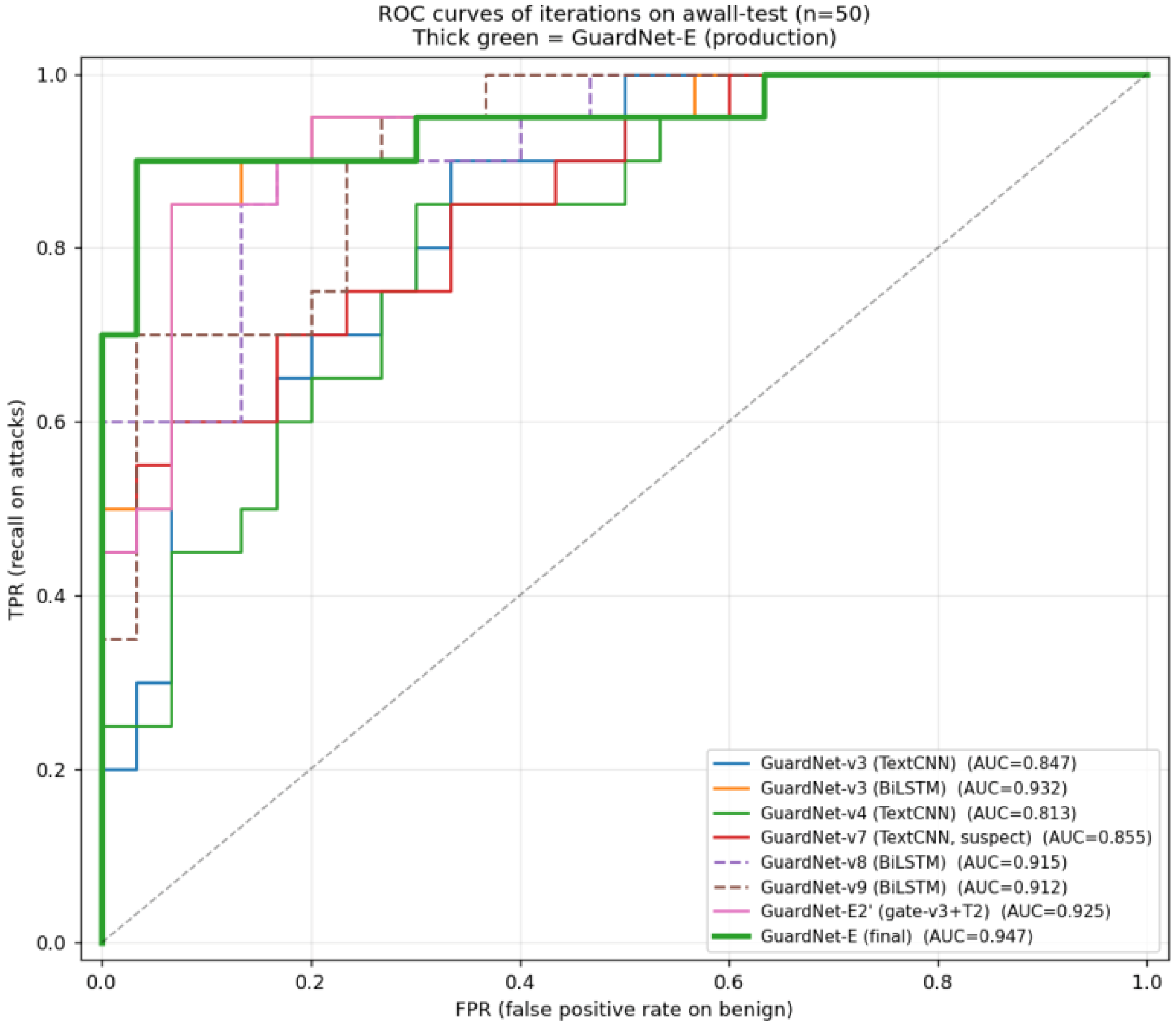

**Figure 2.** ROC curves of GuardNet variants on the awall-test benchmark (n = 50), showing true positive rate (TPR) versus false positive rate (FPR) across architectures.

### 4.2 Effects of Benchmark Contamination: A Reality Check Perspective

One of the main findings of this work is the identification of evidence of memorization in widely used external models. The model protectai-v2, with 184M parameters, achieves perfect performance (F1 = 1.000) on the awall-test benchmark, where overlap with public data exists. However, when evaluated on the unseen JBB-Behaviors pool, its performance collapses to F1 = 0.000, indicating a complete failure to generalize to attacks published after its release. This result highlights the need for segregated and audited test sets to properly measure real-world security performance. Table X presents a comparison of the evaluated models, ranked by the maximum F1 obtained via threshold sweeping. Figure 3 shows the comparison of models on the JBB-Behaviors benchmark, composed of 200 previously unseen examples for all evaluated classifiers. However, it is important to note that this near-perfect performance occurs under a setting with declared partial leakage; in blind evaluations such as JBB-Behaviors, the model still achieves a solid AUC of 0.747, while still outperforming industrial baselines such as protectai-v2.

**Table 4.** Comparison of models evaluated on the blind JBB-Behaviors benchmark (n = 200), ranked by F1_max obtained via threshold sweeping. The table reports the number of parameters, AUROC, benign false positive rate (FPR_b), and harmful content recall (rec_h) at the optimal threshold.

| # | Model | Parameters | F1 max | AUROC | FPR_b@max | rec_h@max |
|---|---|---|---|---|---|---|
| 1 | llm/mistral-7b | 7,2 B | 0,828 | 0,839 | 0,19 | 0,87 |
| 2 | llm/llama3.1-8b | 8,0 B | 0,807 | 0,864 | 0,08 | 0,91 |
| 3 | GuardNet-E (final) | 47 M | 0,714 | 0,747 | 0,53* | 0,85 |
| 4 | deepset-deberta | 184 M | 0,701 | 0,650 | 0,50 | 0,81 |
| 5 | llm/phi3-3.8b | 3,8 B | 0,692 | 0,741 | 0,87 | 1,00 |
| 6 | llm/qwen2.5-7b | 7,6 B | 0,428 | 0,802 | 0,03 | 0,28 |
| 7 | jackhhao-jailbreak | 184 M | 0,137 | 0,574 | 0,09 | 0,08 |
| 8 | protectai-v2 | 184 M | 0,000 | 0,600 | 0,01 | 0,00 |

* Threshold τ = 0.65 was calibrated on a separate hold-out set; threshold sweeping on JBB-Behaviors yields F1_max = 0.714 at τ = 0.80.

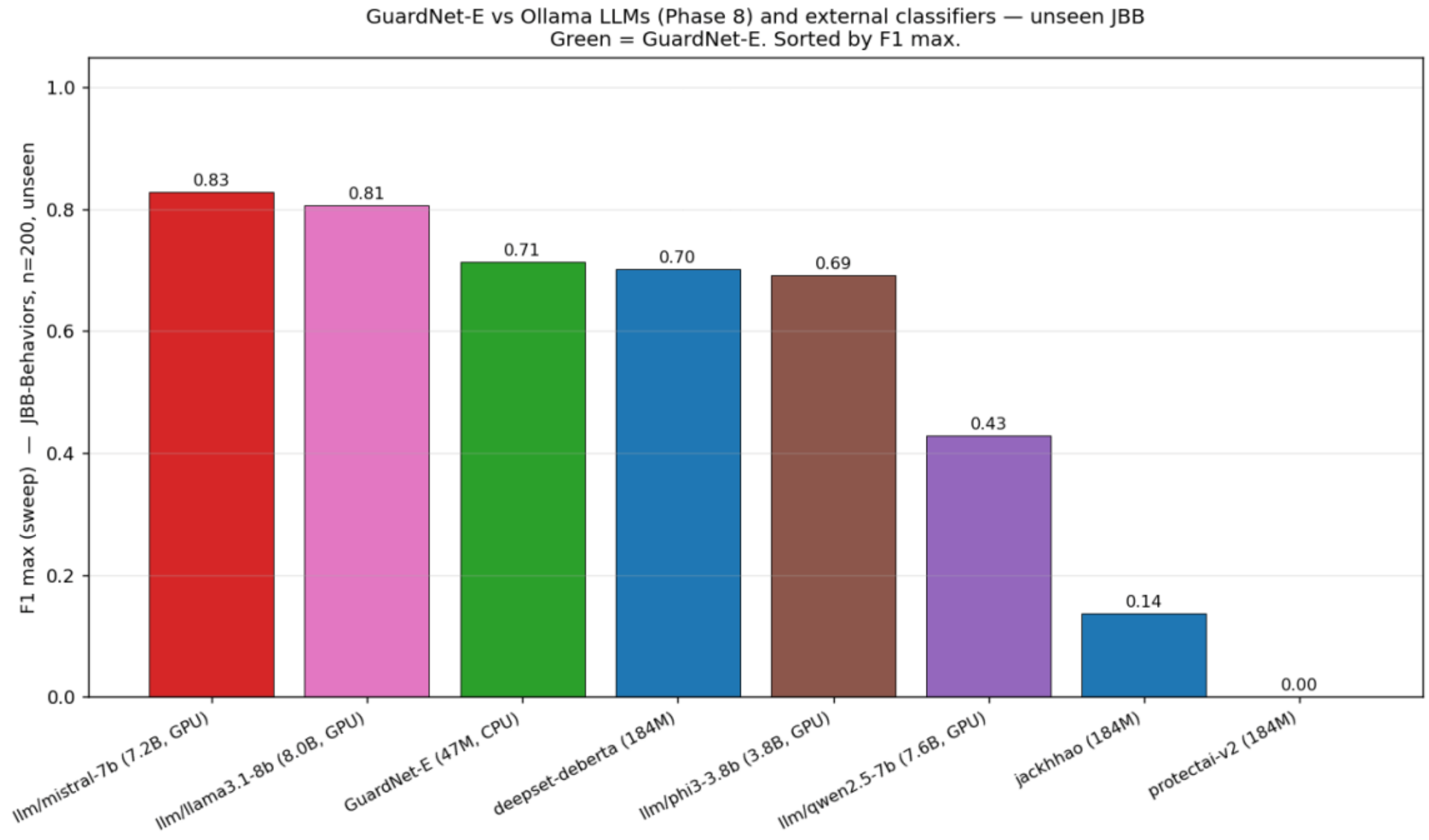

**Figure 3.** Comparison of models on the JBB-Behaviors benchmark (n = 200), composed of previously unseen examples for all evaluated classifiers. LLMs executed via Ollama were accessed through LangChain using structured JSON-schema outputs. The final

score for each model was defined as max ($p_{prompt_injection}$, $p_{jailbreak}$). No failures were observed across the 800 calls performed (4 LLMs × 200 prompts).

### 4.3 Externos Comparative Performance of the Ensemble and External Classifiers

Figure 4 highlights two distinct regimes of behavior between continuous classifiers (curves) and LLMs evaluated as aggregated operating points. For the classifiers, the curves represent the distribution of scores per prompt in the JBB set, enabling a finer-grained analysis of the trade-off between false positive rate and recall across different thresholds. In this setting, GuardNet-E (thick green line) stands out with an AUC of 0.747, consistently outperforming both internal and external baselines, including deepset (0.650), GuardNet-v3 (0.650), protectai-v2 (0.600), and jackhhao (0.574). This result indicates that the internal model not only achieves better global class separation, but also maintains a stable advantage across the entire decision curve, being the only model to simultaneously dominate all three external baselines on a completely unseen data pool.

In generalization tests on the JBB benchmark, GuardNet-E, with 47M parameters, demonstrated empirical superiority over significantly larger architectures. The ensemble outperformed deepset-deberta (184M) in both AUROC (0.747 vs. 0.650) and maximum F1 score. The ROC curves shown in Figure X further support this behavior, indicating that GuardNet-E maintains a better trade-off between recall and false positive rate (FPR), particularly in the low-FPR region. These results suggest that a well-curated ensemble of shallow neural networks can be more robust than large-scale models trained in a more generalist manner, which may be more sensitive to distribution shifts in evaluation pools.

In contrast, the LLMs from the Ollama/LangChain setup do not provide a per-prompt score distribution in this experiment and are instead represented as aggregated operating points defined by FPR_b@max and rec_h@max. This representation reveals a more heterogeneous behavior among models of similar scale. The llama3.1-8b and mistral-7b models occupy the upper-left region of the plot, indicating simultaneously high recall and low false positive rates, which positions them as the best performers in terms of absolute F1 within this group. In contrast, phi3-3.8b is shifted toward the high-FPR region (≈0.87), suggesting over-detection behavior with a strong tendency to classify nearly any input as an attack. Conversely, qwen2.5-7b is concentrated near the origin, reflecting an overly conservative bias, with low false positive rates but also limited recall (≈0.28).

AUROC decouples the choice of threshold, reflecting only the quality of example ranking. In this setting, llama3.1-8b leads in AUROC (0.864), despite achieving a lower maximum F1 than mistral-7b (0.828). GuardNet-E obtains an AUROC of 0.747, placing it below the best-performing LLMs, but still above the external classifiers evaluated. It is worth noting that the LLMs accessed via Ollama appear as single points ("X") in the plot, as they return only binary decisions (fixed operating points). Since no adjustable threshold is available, the reported AUROC for these models should be interpreted as a point estimate, and is not strictly comparable to the full area under continuous ROC curves.

Overall, the analysis reveals that the variability among LLMs with similar architectures is comparable to, and in some cases greater than, the differences between the weakest LLM and the specialized ensemble. This suggests that, for this type of task, calibration and ensemble strategies play a role as important as the choice of backbone architecture, while general-purpose models exhibit extreme behaviors depending on their implicit alignment conFiguretions and sensitivity to adversarial prompts.

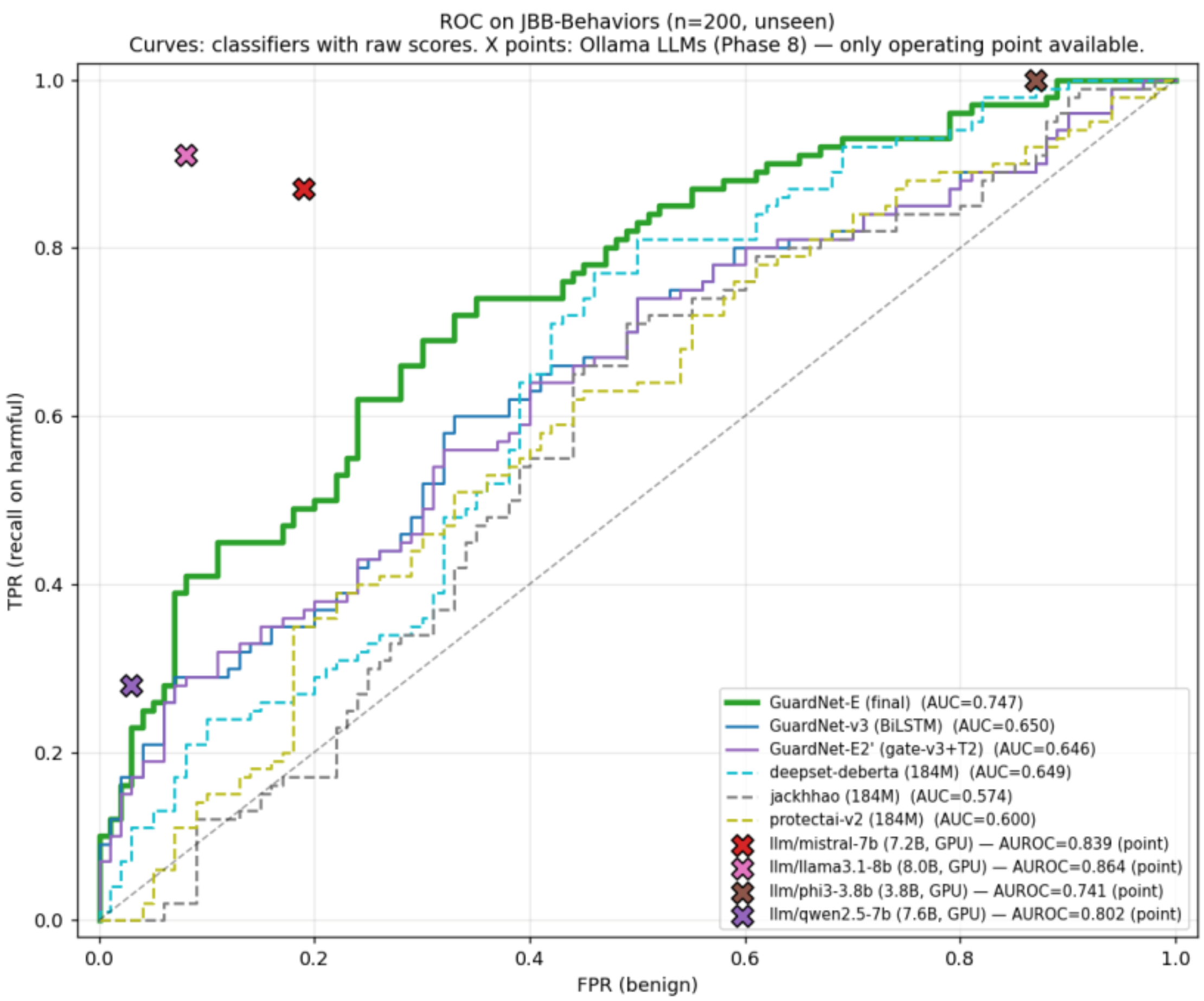

**Figure 4.** Performance on the JBB benchmark. The curves represent continuous classifiers based on per-prompt scores, while the points correspond to LLMs evaluated as aggregated operating points (FPR_b@max, rec_h@max). GuardNet-E (AUC = 0.747) outperforms the other continuous models, while LLMs exhibit greater variability in behavior in terms of recall and false positive rate.

### 4.4 Trade-offs Between GuardNet and LLM-as-a-Judge in Production Settings

Although LLMs such as Mistral-7B and Llama-3.1-8B achieve higher absolute F1 scores (0.828 and 0.807, respectively), GuardNet-E presents structural advantages for production environments, including:

i) Latency: The ensemble operates at approximately 50 ms on CPU, making it roughly 400× faster than LLMs running on the same hardware.
ii) Architectural Immunity: As a discriminative classifier based on BiLSTM networks, without a language modeling head (lm_head), GuardNet does not perform autoregressive generation or token decoding. It operates exclusively as an input-to-label mapper, reducing its exposure to prompt injection attacks targeting instruction-following behavior, since it lacks internal mechanisms for interpreting inputs as generation commands.
iii) Determinism: The model produces consistent probabilistic scores, in contrast to LLMs that may hallucinate justifications and exhibit high variance across models of similar scale.

## 5. Conclusions

The GuardNet project demonstrates that using an ensemble of shallow neural networks (~47M parameters) is a highly effective and efficient alternative to large language models (LLMs) for detecting prompt injection and jailbreak attacks. The empirical results support the premise that diversity of adversarial sources is more important than parameter scale for generalization in security tasks. By outperforming significantly larger models on unseen datasets, GuardNet-E suggests that rigorous data curation, combined with anti-dominance sampling strategies and threshold calibration, enables the construction of robust guardrails with minimal latency (~50 ms on CPU), deterministic behavior, and lightweight deployment suitable for production environments.

However, the analysis also shows that performance is not uniform across all evaluation regimes, with larger LLMs still exhibiting advantages in certain settings in terms of F1 and AUROC on the blind benchmark. This reinforces that the superiority of the proposed method is context-dependent, particularly when constraints on cost, latency, and architectural control are prioritized.

In summary, GuardNet does not aim to replace general-purpose LLMs, but rather to provide a specialized and efficient alternative for adversarial attack detection in security pipelines, particularly in production contexts where predictability and operational efficiency are critical factors.

## 6. Perspectives for Future Work

The next phase in the evolution of Awall involves the development of a proprietary safety classifier, as a successor to the current generation, specifically designed for the security gateway domain. This evolution is guided by three objectives: (i) consolidating intellectual property over the architecture, dataset, taxonomy, and evaluation suite; (ii) enabling CPU-based inference, thereby expanding support for self-hosted customers without dedicated GPU infrastructure; and (iii) aligning the taxonomy with the MLCommons AILuminate v1.0 standard, enabling public comparability against official benchmarks and paving the way for academic publication.